\documentclass[letterpaper, 10 pt, conference]{ieeeconf}  
\IEEEoverridecommandlockouts                              
\usepackage{amsmath}
\usepackage{amssymb}
\usepackage{booktabs}
\usepackage{cite}
\usepackage{comment}
\usepackage{float}
\usepackage{graphics} 
\usepackage{graphicx}
\usepackage{multirow}
\usepackage{subcaption}

\PassOptionsToPackage{hyphens}{url}\usepackage[breaklinks=true,hidelinks=true,colorlinks=true,citecolor=blue,linkcolor=blue]{hyperref}

\graphicspath{{figures/}}

\makeatletter
\renewcommand{\fnum@figure}{Fig. \thefigure}
\makeatother

\makeatletter
\renewcommand{\fnum@table}{Table \thetable}
\makeatother

\title{\LARGE \textbf{
Simulation-Based Multi-Fillet Evaluation\\ of Woody Breast Poultry Fillets 
}}

\author{Chirantan Sen Mukherjee$^{1}$, Seung-Chul Yoon$^{2}$, and William J. Beksi$^{1}$
\thanks{$^{1}$
The authors are with the Department of Computer Science and Engineering, 
The University of Texas at Arlington, Arlington, TX, USA. 
Emails:
cxs6305@mavs.uta.edu,
william.beksi@uta.edu.
}%
\thanks{$^{2}$
The author is with the Quality and Safety Assessment Research Unit, 
U.S. National Poultry Research Center, 
USDA Agricultural Research Service, Athens, GA, USA.
Email:
seungchul.yoon@usda.gov.
}}

\begin{document}

\maketitle
\pagestyle{plain}

\begin{abstract}
\label{sec:abstract}
Woody breast (WB) is a myopathy in modern broiler chickens that causes the
breast muscle to become unusually stiff and fibrous, leading to decreased meat
quality and significant economic losses. State-of-the-art automated WB detection
relies on a side-view imaging system to analyze the bending behavior of a single
fillet as it falls off a conveyor belt. While highly accurate, this approach is
constrained by its single-fillet field of view, creating throughput bottlenecks
on commercial processing lines. In this paper, we address this limitation via a
novel multi-fillet detection architecture utilizing a top-down camera
configuration. To validate our approach, we first develop a high-fidelity
digital twin of an industrial conveyor system. Next, we synthesize a diverse
dataset of 3D fillet meshes and model their viscoelastic bending dynamics using
a physics-based simulation engine. Lastly, a continuous 2D shape deformation
score is extracted from the top-down perspective as the simulated fillets
traverse the roller precipice. Experimental results demonstrate that the
top-down shape score effectively captures the contour changes of the fillets as
it bends, providing a robust and scalable alternative to a side-view imaging
system for simultaneous multi-fillet WB evaluation.
\end{abstract}

\section{Introduction}
\label{sec:introduction}
Chicken breast meat is an affordable, lean protein source, favored by consumers
for its low fat and high protein content~\cite{praud2020molecular}. The global
increase and demand for poultry products has driven the industry to prioritize
rapid growth for maximum meat yield in broiler chickens. To increase bird size,
producers have extended growth periods and performance. However, this intensive
approach has resulted in myopathy-associated leg disorders, gait impairment,
cardiovascular diseases, and high mortality
rates~\cite{siegel1997genetic,caldas2020review,zhang2021broiler}. Furthermore,
the increased growth rate in modern broilers has contributed to the prevalence
of woody breast (WB), a myopathy (muscle disease) that adversely affects the
quality of the meat.

The WB condition is characterized by an abnormally hard and rigid texture of the
bird's pectoral muscle. There are visual cues such as bulging toward the caudal
end of the fillet and pale discoloration and surface
hemorrhages~\cite{sihvo2014myodegeneration}. Nevertheless, relying on these
signals is problematic since they are not consistently present in all affected
fillets. This makes it difficult to accurately and reliably identify WB without
physical palpation or objective instrumental assessments such as histological
analysis.

The poultry industry needs an objective, rapid detection methodology capable of
identifying WB fillets at commercial line speeds. A 2016 study estimated that WB
costs the U.S. poultry industry over \$200 million in annual losses. This is
largely because affected fillets are downgraded from premium fillets and
diverted to further-processing lines, such as ground-meat products, where they
require additional handling and yield lower profit
margins~\cite{gee2016poultry,kuttappan2016white}. In addition, brand damage
further impacts the poultry industry economically since the abnormal textural
profile of WB fillets decreases consumer satisfaction.

\begin{figure}[t]
\centering
\includegraphics[width=\linewidth]{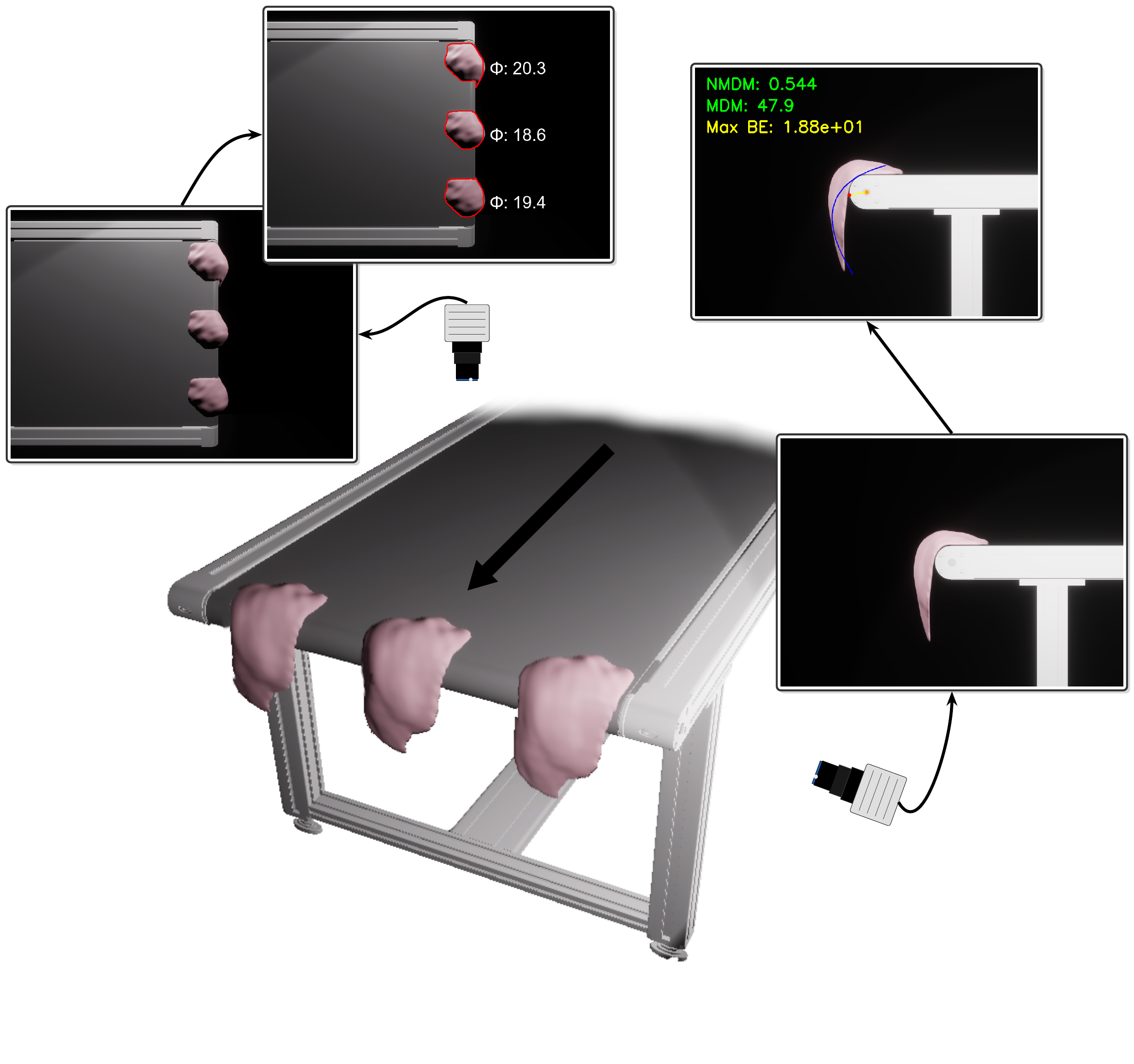} 
\caption{An overview of our WB simulation framework. The system simulates an
industrial conveyor system transporting chicken breast fillets. While the
side-view camera provides baseline metrics, we strictly utilize only the
top-down camera for multi-fillet WB evaluation.}
\label{fig:overview}
\end{figure}

Given the scale of this issue, including its prevalence in commercial production
and associated economic losses, it is critical to reliably and accurately detect
and distinguish WB fillets from normal fillets. To address this challenge, we
introduce a framework that detects multiple fillets on a simulated industrial
processing line, Fig.~\ref{fig:overview}. In summary, the main contributions of
our work are as follows.
\begin{itemize}
  \item We propose a high-fidelity simulation environment of an industrial
  conveyor system with deformable fillet models.
  \item We release a public synthetic 3D fillet dataset and the accompanying
  simulation source code to the research community.
  \item We develop an approach that simultaneously tracks multiple fillets and
  differentiates between normal and WB fillets by exclusively employing a
  top-down camera perspective.
\end{itemize}
The source code, dataset, and multimedia associated with this project can be
found at
\href{https://robotic-vision-lab.github.io/woody-breast-sim}{https://robotic-vision-lab.github.io/woody-breast-sim}.

The remainder of this paper is organized as follows. Sec.~\ref{sec:related_work}
reviews prior research on WB detection and synthetic data generation. In
Sec.~\ref{sec:methodology}, we explain the development of our physics-based
simulation environment, the synthetic dataset creation, and the multi-fillet
tracking algorithms. Sec.~\ref{sec:evaluation} presents the experimental setup
and analyzes the results of both the side-view and top-down camera bending
evaluations. Finally, Sec.~\ref{sec:conclusion_and_future_work} concludes the
paper and outlines future work.

\section{Related Work}
\label{sec:related_work}
\subsection{Woody Breast Detection}
Prior research has extensively characterized WB using both biological and
imaging approaches. For example, offline computer vision
techniques~\cite{geronimo2019computer}, near-infrared
spectroscopy~\cite{wold2019near}, and optical coherence tomography
(OCT)~\cite{ekramirad2024nondestructive,pallerla2024neural} can successfully
differentiate WB from normal tissue based on features such as decreased protein
content~\cite{wold2017rapid} and sub-surface
microstructures~\cite{yoon2016toward}. Nonetheless, these methods are largely
ill-suited for online industrial detection. Histological and gene expression
studies~\cite{velleman2015histopathologic,soglia2017superficial} are inherently
destructive and time consuming. Furthermore, high-resolution imaging systems
like OCT require static positioning or slow scanning speeds, which creates major
bottlenecks. This makes them impractical for commercial processing lines where
multiple fillets must be assessed simultaneously at high speeds. 

Compared to visual examination, physical scoring for WB is more complex than
other myopathies due to the difficulty of determining severity via tactile
evaluation~\cite{kuttappan2016white}. Yoon et al.~\cite{yoon2022development}
addressed this gap by measuring the biomechanical properties of a fillet. In
particular, side-view imaging was used to quantify the degree in which a fillet
bends under gravity as it falls from the conveyor belt. This direct measurement
of rigidity provides a robust indicator of the WB condition and achieved a
classification accuracy of over 95\%. However, the approach was validated only
for individual fillets moving in a single file. Its primary limitation is the
inability to detect and track multiple fillets simultaneously, a requirement for
high-throughput industrial processing. Our work builds upon this biomechanical
principle, but adapts it for multi-fillet detection.

\subsection{Synthetic Data Generation}
Generating lifelike training data for raw agricultural products is extremely
challenging due to the high variability of biological shapes and the scarcity of
large, annotated 3D datasets. Although generative adversarial networks (GANs)
have shown immense potential in synthesizing realistic-looking data, standard
architectures including StyleGAN2~\cite{karras2020analyzing} require thousands
of images to converge. This requirement is impractical for specialized domains
such as poultry processing. Few-shot GAN adaptation can address this need for
custom data by fine-tuning a generator pretrained on a large data source to a
target domain using limited samples. For instance, early approaches such as
TransferGAN~\cite{wang2018transferring} simply fine-tuned the initialized
weights of the pretrained model. 

To improve relative distances between instances in the source and target
domains, Ojha et al.~\cite{ojha2021few} utilized cross-domain correspondence
(CDC). However, this technique struggles when the domains differ significantly
in geometry. Specifically, CDC relies on the assumption of a structural
correspondence between the domains. This assumption breaks down in our scenario
where the source (architectural buildings) shares no semantic structure with the
target (poultry fillets). To overcome this, we use smoothness similarity
regularization (SSR)~\cite{sushko2023smoothness}. Unlike CDC, SSR does not
enforce instance-level matching. Instead, it transfers the inherent latent space
smoothness of the pretrained model, ensuring that the rate of change in the
generated geometry remains stable. This allows for the creation of a diverse
group of synthetic 3D fillet depth maps from a small set of real-world scans
without the overfitting and mode collapse observed in prior approaches.

\section{Methodology}
\label{sec:methodology}
\begin{figure*}
\centering
\includegraphics[width=\linewidth]{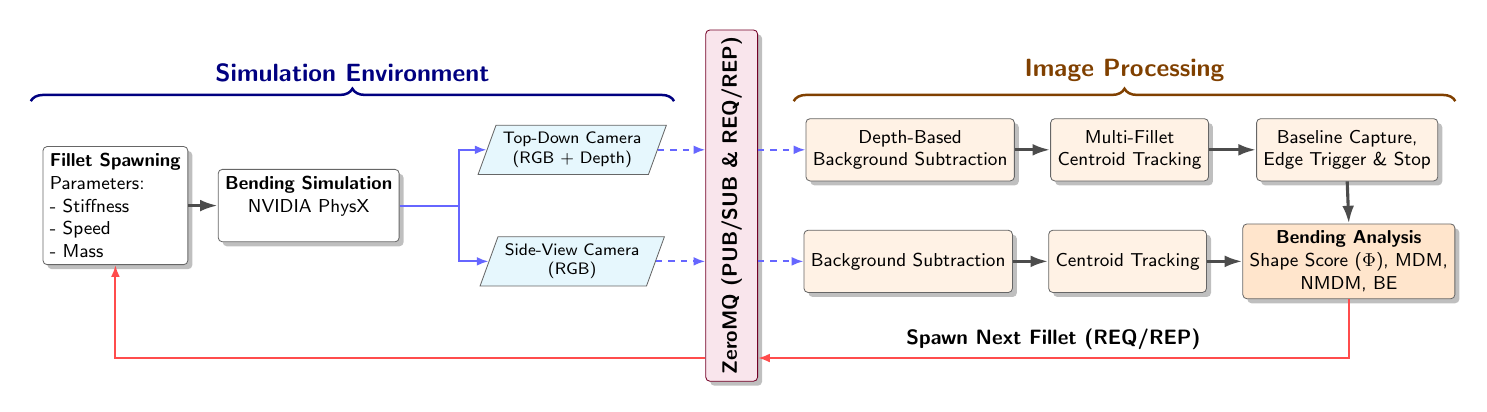}
\caption{The system architecture and data flow pipeline for our automated WB
inspection simulation. Within the simulation environment (left), fillet meshes
are spawned with specific stiffness, speed, and mass parameters utilizing NVIDIA
PhysX to simulate realistic bending deformation over a conveyor belt roller.
During the image processing stage (right) the image and depth streams from the
simulation are analyzed to isolate the fillets and determine their bending
scores.}
\label{fig:pipeline}
\end{figure*}

We propose a bespoke simulation framework that provides a faithful environment
for testing online WB detection algorithms. To achieve this, we first synthesize
a large-scale dataset of 3D fillet meshes from a small set of real-world scans
using a generative model. Then, we develop WB detection strategies within the
physics-based simulator. An overview of the processing pipeline is provided in
Fig.~\ref{fig:pipeline}.

\subsection{Real-World Data Collection}
We collected 40 high-quality 3D point cloud scans of broiler fillets using an
RGBD camera. Nevertheless, training a 3D generative model directly on such a
limited dataset is highly prone to overfitting and mode collapse. To overcome
this dilemma, we convert the 3D point clouds into 2D depth maps. This
transformation allows us to leverage a pretrained 2D generative architecture
while decoupling the fillet's underlying morphological shape from its absolute
physical scale. To ensure that the generative model learns accurate topological
variance without being biased by specific sizes or poses, we implement a
post-processing pipeline to standardize the data prior to projection.

\subsection{Real-World Data Post-Processing}
After collecting real fillet scans, we post-process the data as follows. First,
we employ principal component analysis to align the primary axes of each fillet
in the same direction to remove rotational variance. Each aligned point cloud is
scaled such that its maximum spatial extent fits perfectly within a unit sphere.
Next, the normalized point clouds are projected onto a 2D grid to create depth
maps. To assure that the generative model will learn accurate relative thickness
variations between different samples, the pixel intensities of the depth maps
are normalized against a maximum height calculated across the entire dataset.

\subsection{Synthetic Dataset Details} 
Relying solely on a limited set of 40 fillet scans is insufficient for
validating a robust WB detection system. Biological products exhibit large
morphological variance, and a WB classifier validated on such a small sample
size would not generalize to the shape and size diversity experienced in an
industrial poultry processing line. We address this problem by using the
preprocessed depth maps as a seed dataset for a StyleGAN2-based architecture
modified with SSR to produce a diverse set of depth maps. The generator is
pretrained on the LSUN-Churches~\cite{yu2015lsun} dataset with dimensions
$256\times 256$. Our dataset consists of 1,000 synthetic 3D meshes that capture
the natural variance of poultry fillets. These 3D models are generated by first
synthesizing 2D depth maps using the modified GAN architecture, which are
subsequently reconstructed into volumetric 3D meshes.

\subsection{Synthetic Data Augmentation}
Applying SSR results in the gradients of the target generator's feature maps
aligning with those of the source generator. This change effectively transfers
the latent space smoothness of the pretrained model to the fillet domain,
ensuring that interpolations in the latent space result in continuous geometric
deformations rather than discrete jumps. The loss function is defined as
\begin{equation}
  \mathcal{L}_{\text{SS}} = \lambda_{\text{SS}} \cdot \mathbb{E}_{(z,y) \sim \mathcal{N}(0,1)} || \nabla_z \langle G_{s}^l(z), y \rangle - \nabla_z \langle G_{t}^l(z), y \rangle ||_2,
\end{equation}
where $\lambda_\text{SS}$ is the smoothness similarity, $G_s$ and $G_t$ are the
source and target generators, and $y$ is a random Gaussian projection vector. We
set $\lambda_{\text{SS}} = 0.2$ due to the structural dissimilarity between the
source and the target domain. This relaxation reduces the influence of the
pretrained model's geometric priors so that the generated fillet depth maps look
realistic. Unlike the standard StyleGAN2 discriminator, which outputs a single
validity score, the SSR discriminator includes auxiliary heads ($1\times 1$
convolutions) at every resolution block. This supervision forces the model to
maintain global shape consistency while generating local surface details and
preventing overfitting. 

To utilize the 2D depth maps generated by the StyleGAN2-SSR model within our
physics-based simulation, we reconstruct them into 3D volumetric meshes. During
reconstruction, we generate vertices by mapping the 2D pixel coordinates and
intensity values into 3D space. The maximum length and thickness of each
generated fillet is then scaled to match the physical dimension ranges reported
by Yoon et al.~\cite{yoon2022development}. A flat bottom surface is generated
for each fillet and the boundary edges are stitched to create a closed,
watertight mesh. 

\subsection{Simulation Environment Development}
The simulation environment was created using Unity's High Definition Render
Pipeline~\cite{unity}. The virtual layout is similar to the side-view imaging
setup described by Yoon et al.~\cite{yoon2022development}. It features a
variable-speed conveyor belt, a camera facing the roller axle of the conveyor,
and a top-down camera with a nadir view of the conveyor belt. 

\begin{figure}
\centering
\begin{subfigure}{0.237\textwidth}
  \centering
  \includegraphics[width=\linewidth]{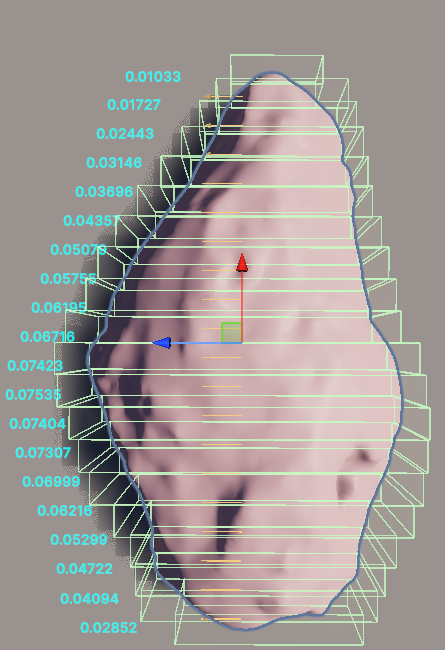}
  \caption{}
  \label{f1s1}
\end{subfigure}
\begin{subfigure}{0.2\textwidth}
  \centering
  \includegraphics[width=\linewidth]{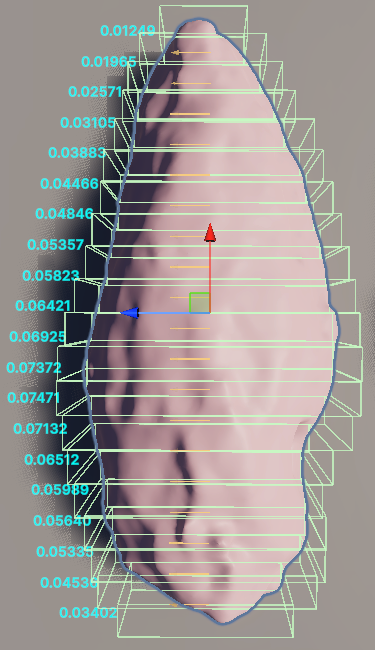}
  \caption{}
  \label{f1s2}
\end{subfigure}
\caption{The longitudinal mass distribution profiles of two synthetic fillet
models with a mass of 1 kg.}
\label{fig:fillet_mass}
\end{figure}

To simulate the nonrigid behavior of the fillets, we model each fillet mesh as a
discretized chain of $N$ rigid bodies connected by torsional springs. The motion
of each segment $i$ in the mesh relative to its neighbor is governed by the
rotational equation of motion for a damped harmonic oscillator,
\begin{equation}
  I_i \ddot{\theta}_i + c \dot{\theta}_i + k \theta_i = \tau_{\text{ext}},
  \label{eq:harmonic_oscillator}
\end{equation}
where $\theta_i$ is the angular displacement and $\tau_{\text{ext}}$ represents
the torque incurred from gravity. The system parameters rely on inertial
properties ($I_i$), damping ($c$), and stiffness ($k$). The mass distribution is
nonuniform and derived directly from the voxel volume of the generated meshes.
The mesh is sliced along the longitudinal axis, and a mass $m_i$ is assigned to
each rigid body segment proportional to its local cross-sectional area
(Fig.~\ref{fig:fillet_mass}). This ensures that the center of gravity shifts
realistically during motion. A viscoelastic damping term $\tau = -c\dot{\theta}$
is applied to mitigate numerical jitter and simulate internal tissue friction.
Finally, the stiffness coefficient $k$ serves as the primary variable for
simulating WB myopathy. We model normal fillets as highly compliant chains ($k
\approx 0$) that drape continuously over the roller, whereas WB fillets are
modeled with high-torsional stiffness ($k \gg 0$) to resist curvature. 

A challenge in simulating an industrial conveyor system is the standard friction
models, which can lead to micro-slippage and inconsistent travel speeds. We
tackle this issue by implementing a custom kinematic velocity drive controller.
Instead of relying on Coulomb friction, our controller calculates the precise
force required to lock the fillet's contact velocity to the target line speed
($V_{\text{target}}$) at every simulation step,
\begin{equation}
  F_{\text{drive}} = \frac{m(V_{\text{target}} - V_{\text{current}})}{\Delta t},
\end{equation}
where $V_{\text{current}}$ is the current contact velocity of the fillet
segment. This drive force is conditionally applied based on the surface normal
of the contact point. To prevent the conveyor system from artificially driving
the fillet down the curve of the roller, we monitor the dot product between the
contact normal and the global up vector. If downward deviation is detected, then
the drive force is decoupled for that segment ensuring that the fillet enters a
gravity-dominant trajectory immediately upon reaching the roller edge.

\subsection{Synthetic Data Evaluation Strategy}
To ensure the synthetic depth maps accurately reflect the geometric properties 
of real fillets, we employ an evaluation strategy that focuses on both 
distribution quality and sample diversity. Given the limited size of the training 
set, metrics such as the Fr\'{e}chet inception distance~\cite{heusel2017gans} 
can be unstable. Therefore, we utilize the kernel inception distance 
(KID)~\cite{binkowski2018demystifying} and learned perceptual image patch similarity
(LPIPS)~\cite{zhang2018unreasonable}. However, the problem with the fillet depth
maps when calculating LPIPS scores is the dominance of the black background.
LPIPS will yield misleadingly high similarity scores since the majority of
pixels in both real and synthetic images are zero-value background. We solve
this problem by cropping the images before evaluation. In detail, both real and
generated images are dynamically cropped to the bounding box of the fillets and
resized to $256\times 256$ prior to feature extraction. This forces the metric
to evaluate the shape of the fillet rather than the consistency of the
background. 

\subsection{Fillet Bending Evaluation Strategy}
As a fillet falls from the conveyor belt, its rigidity is evaluated continuously
using the minimum distance measure (MDM), normalized minimum distance measure
(NMDM), and bending energy (BE). These metrics are all determined from the
side-view camera stream. For each frame $t$, the 2D contour of the fillet is
segmented, and its spatial centroid $C_t = (x_c, y_c)$ is calculated using the
image moments. Then, the MDM is computed to quantify the absolute spatial
displacement of the fillet's center of mass relative to the conveyor belt
roller, i.e.,
\begin{equation}
  \text{MDM}_t = \| C_t - P_{\text{roller}} \| = \sqrt{(x_c - x_r)^2 + (y_c - y_r)^2},
  \label{eq:mdm}
\end{equation}
where $P_{\text{roller}} = (x_r, y_r)$ is the static pixel coordinates of the
roller axle. A highly rigid fillet (WB) will project further horizontally into
3D space, yielding a larger MDM. Conversely, a flexible fillet will yield to
gravity and drape closer to the roller axle. 

Since broiler fillets vary naturally in size, the raw distance must be
normalized. To this end, we adopt the NMDM as our primary shape descriptor. The
NMDM normalizes the minimum observed distance against the fillet's thickness.
Concretely,
\begin{equation}
  \text{NMDM} = \frac{\min_{t} (\text{MDM}_t)}{R + h_{t}},
  \label{eq:nmdm}
\end{equation}
where $R$ is the radius of the conveyor belt roller (in pixels), and $h_{t}$ is
the vertical distance from the fillet's centroid to its top surface, captured on
the flat portion of the conveyor belt.

The BE describes the total structural deformation (i.e., curvature) of a fillet.
To compute the BE, we extract the 1D medial axis (skeleton) of the 2D contour
using a morphological thinning algorithm. The skeleton coordinates are then
approximated using a least-squares second-degree polynomial fit, $y(x) = ax^2 +
bx + c$. The local curvature at any point along this polynomial is defined as
\begin{equation}
  \kappa = \frac{2|a|}{(1 + (2ax + b)^2)^{\frac{3}{2}}}.
  \label{eq:curvature}
\end{equation}
The BE for a given frame is calculated by integrating the squared curvature
across the skeleton. To guarantee that the metric is scale invariant, the sum is
normalized by the squared contour perimeter $P^2$ and scaled by the skeleton arc
length $L$, i.e.,
\begin{equation}
  \text{BE}_t = \frac{P^2}{L} \sum \kappa^2.
  \label{eq:be}
\end{equation}

\subsection{Multi-Fillet Woody Breast Evaluation}
While the side-view camera provides ground-truth bending metrics for individual
fillets, our main contribution is multi-fillet WB evaluation. To achieve this,
our simulation utilizes a top-down RGBD camera to isolate and track multiple
fillets simultaneously as follows. We first apply depth-based background
subtraction followed by watershed segmentation to the camera data stream. This
allows us to accurately delineate individual fillets even when they are
positioned in close proximity on the conveyor belt. The fillets are then
continuously tracked across frames by minimizing the Euclidean distance between
their geometric centroids.

Our multi-fillet evaluation approach relies on analyzing the geometric change of
the fillets as they transition over the conveyor belt roller. Before a fillet
reaches the roller edge, its 2D contour ($H_{\text{ref}}$), rotated bounding box
aspect ratio (AR), and baseline solidity ($S_{\text{ref}}$) are recorded.
Industrial processing lines utilize wide conveyor belts. Therefore, a single
top-down camera introduces perspective distortion, i.e., a fillet positioned at
the edge of the belt will yield a slightly different 2D contour than the exact
same fillet positioned in the center lane. To account for this spatial variance,
we spawn identical fillet meshes across multiple lanes during simulation to
ensure that the baseline metrics are robust against the camera's perspective
shift. 

Since the tail end of a modern broiler breast exhibits varying degrees of 
compliance that can introduce significant noise in the deformation analysis 
regardless of the WB condition, its inclusion is problematic. To address this 
problem, we dynamically detect the tail end of each fillet and amputate it 
from the active contour. Specifically, we measure the transverse width along 
the longitudinal axis of the fillet and amputate the tail end where the width 
drops below 30\% of the maximum body width. This implementation restricts the 
structural analysis strictly to the main body of the muscle where rigidity is 
most prominent. As the amputated body advances and its centroid crosses the 
conveyor belt roller, its top-down footprint changes. To quantify this 
deformation, we calculate a shape score as follows. First, we compute the 
solidity of the visible contour, 
\begin{equation}
  S = \frac{A_c}{A_h},
\end{equation}
where $A_c$ is the contour area and $A_h$ is the area of its convex hull. Next,
we extract the Hu moments of the visible contour and compare them to the moments
of the reference body contour. We quantify the geometric distortion using a
normalized matching distance $D(H_{\text{ref}}, H_{\text{cur}})$ based on the
reciprocal logarithm of the Hu moments. To make certain that this metric is
robust across naturally varying fillet proportions, the distance is normalized
by the AR of the reference fillet. The final shape score combines this
normalized moment score with the change in solidity,
\begin{equation}
  \Phi = w_1 \left( \frac{D(H_{\text{ref}}, H_{\text{cur}})}{\text{AR}} \right) + w_2 |S_{\text{ref}} - S_{\text{cur}}|,
\end{equation}
where $S_{\text{cur}}$ is the solidity of the contour in the current frame and
$w_1, w_2$ are weighted coefficients. In our experiments, both $w_1$ and $w_2$
are set to 100 to equally scale the normalized moment and solidity differences
for readability. The shape score is continuously updated as the fillet advances
over the edge of the belt. To account for the deformation of the fillet shifting
its center of mass point during the drop, we lock the spatial offset between the
main body's centroid and the rigid trailing edge of the fillet while it is still
flat. We then project this virtual reference centroid forward based on the
tail's position. We terminate the evaluation window for the maximum observed
shape score at the exact moment this virtual reference centroid crosses the
conveyor belt edge.

\section{Evaluation}
\label{sec:evaluation}
To validate the fidelity of our simulation framework, we conducted drop tests
designed to replicate the biomechanical analysis performed by Yoon et
al.~\cite{yoon2022development}. The objective was to determine how the simulated
stiffness parameter, fillet mass, and conveyor line speeds correlate with the
physical bending profiles of the chicken fillet meshes. In the following
subsections we detail the results of the evaluation.

\subsection{Experimental Setup} 
We utilized 1,000 synthetic fillet meshes whose depth maps were generated using
StyleGANv2-SSR model. The model was trained using 360,000 images shown to the
discriminator with a batch size of 4. We disabled standard augmentations to
preserve the specific geometric orientation of the fillets required for the
simulation. All experiments and simulations were conducted on a workstation
equipped with an AMD Ryzen 7 7800X3D processor, 32 GB of RAM, and an NVIDIA
GeForce RTX 4070 Ti GPU with 16 GB of VRAM.

\subsection{Synthetic Data Quality Evaluation}
\begin{table}
\begin{center}
\begin{tabular}{lcl}
\hline
\textbf{Metric} & \textbf{Value} & \textbf{Interpretation} \\
\hline
KID score ($\downarrow$)       & 0.081 & High distribution similarity \\
Real diversity (LPIPS)         & 0.305 & Baseline geometric variance \\
Synthetic diversity (LPIPS)    & 0.289 & Generated geometric variance \\
$\text{Real}/\text{synthetic}$ & 95\%  &  \\
\hline
\end{tabular}
\caption{SSR evaluation using StyleGANv2.}
\label{tab:ssr_eval}
\end{center}
\end{table}

We evaluated the model using the generated depth maps against the 40 real
training samples. The results are summarized in Table~\ref{tab:ssr_eval}. A KID
score of 0.081 was achieved. This indicates that the feature distribution of the
synthetic data is statistically close to the real data, despite the few-shot
training regime. The real dataset had a LPIPS of 0.305, while the synthetic
dataset achieved a LPIPS of 0.289. These results show that the generator
captures approximately 95\% of the geometric variance present in the real fillet
data without suffering from mode collapse. 

\begin{figure*}
\centering
\includegraphics[width=\linewidth]{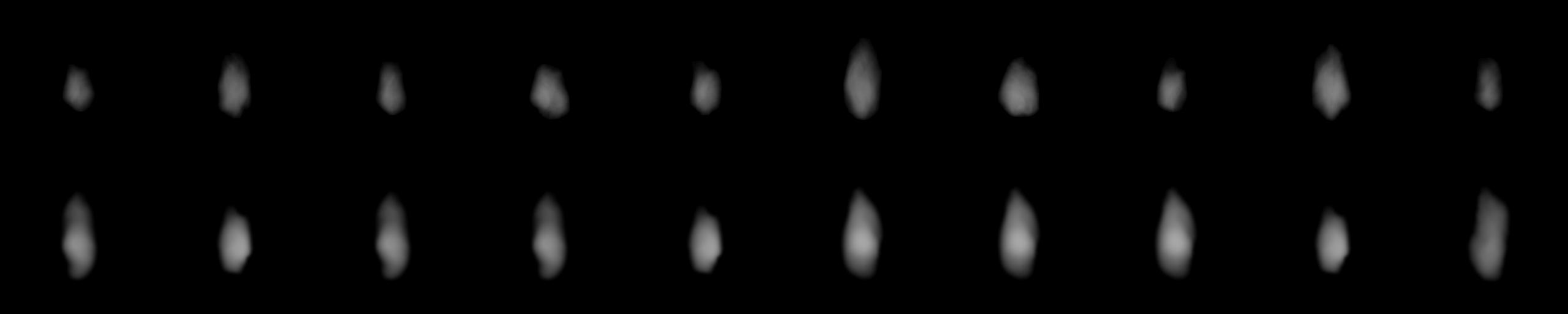}
\caption{A nearest neighbor analysis. The top row shows the generated synthetic
depth maps. In the bottom row, the closest corresponding real depth map from the
training set (determined via LPIPS minimization) is displayed. The visual
differences between the pairs demonstrate that the model generates novel samples
and does not simply memorize the training data.}
\label{fig:nn_analysis}
\end{figure*}

Local variation was measured by clustering synthetic samples by their nearest
real neighbor. The mean pairwise distance within these clusters was 0.277. This
value is close to the global average and indicates that the model produces
significant variations, even when mimicking a specific training example. To
verify the absence of memorization, we conducted a nearest neighbor analysis
using LPIPS as the distance metric. Fig.~\ref{fig:nn_analysis} presents pairs of
synthetic samples and their closest real counterparts from the training set. A
visual inspection confirms that while the synthetic samples share general
structural characteristics with their nearest neighbors, they exhibit distinct
local features and boundary variations. The results show that the model learns
to generalize beyond the 40 training samples rather than simply memorizing them.

\subsection{Side-View Camera Results}
We tested two line speeds on the conveyor belt: 100 and 130 feet per minute
(FPM). These values reflect the standard line speeds of commercial poultry
processing plants in the U.S. A side-view RGB camera captured the bending
profile of the fillets at 200 frames per second. To accurately replicate the
bending of the chicken fillets, the total simulated fillet mass ($M$) and the
torsional stiffness coefficient ($k$) must be treated as a coupled system. Yoon
et al.~\cite{yoon2022development} reported an average mass of 0.46 kg for the
broiler fillets utilized in their physical drop tests, which served as our
initial baseline. Nonetheless, as detailed in Sec.~\ref{sec:methodology}, our
simulation does not assign a global mass to a singular physics collider.
Instead, the total mass is distributed across a discretized chain of $N$
segments that make up the fillet mesh, with each segment assigned a localized
mass $m_i$ proportional to its volume.

In our spring-damper model, the draping of the fillet over the roller is driven
by the localized gravitational torque acting on each segment over time. At high
conveyor speeds of 100 and 130 FPM, the transit time over the roller is reduced
to a fraction of a second. Within this brief window, dividing a total mass of
0.46 kg across the segmented chain results in localized tip masses ($m_i$) that
are too small to generate sufficient downward torque. Therefore, we used the
total mass of the fillet as a tunable parameter to force the fillet mesh to
deform at speed. Since mass and stiffness are directly proportional in
determining physical deformation, we calibrated the simulated mass to match the
biological bounds established by Yoon et al.~\cite{yoon2022development}. In
detail, it was reported that across line speeds ranging from 10 to 100 FPM, the
average physical NMDM for normal fillets was $0.41 \pm 0.10$, while severe WB
fillets measured $0.66 \pm 0.08$, representing a separation range of
$\Delta\text{NMDM} \approx 0.25$.

\begin{table}
\centering
\footnotesize 
\setlength{\tabcolsep}{2.5pt} 
\renewcommand{\arraystretch}{1.0} 
\begin{tabular}{ll ccc}
  \toprule
  \textbf{Line Speed} & \textbf{Condition} & \textbf{NMDM} & \textbf{MDM} & \textbf{BE} \\
  \midrule
        
  \multicolumn{5}{c}{\textbf{Biological Ground Truth (Yoon et al.~\cite{yoon2022development})}} \\
  \midrule
  \multirow{3}{*}{Avg. (10-100 FPM)} 
  & Normal      & 0.41 $\pm$ 0.10 & 41 $\pm$ 11 & 35 $\pm$ 20 \\
  & Moderate WB & 0.63 $\pm$ 0.05 & 63 $\pm$ 5  & 9 $\pm$ 5  \\
  & Severe WB   & 0.66 $\pm$ 0.08 & 67 $\pm$ 7  & 7 $\pm$ 7  \\
        
  \midrule
  \multicolumn{5}{c}{\textbf{Simulated Digital Twin ($M=100$)}} \\
  \midrule
        
  \multirow{5}{*}{100 FPM} 
  & $k = 0$ (normal)   & 0.520 & 45.8 & 37.6 \\
  & $k = 10$           & 0.601 & 52.6 & 25.1 \\
  & $k = 100$ (severe) & 0.783 & 68.3 & 19.3 \\
  & $k = 200$          & 0.824 & 71.8 & 10.0 \\
  & $k = 400$          & 0.848 & 74.0 & 10.5 \\
  \midrule
        
  \multirow{5}{*}{130 FPM} 
  & $k = 0$ (normal)   & 0.675 & 59.0 & 27.8 \\
  & $k = 10$           & 0.716 & 62.6 & 17.4 \\
  & $k = 100$ (severe) & 0.820 & 71.5 & 12.9 \\
  & $k = 200$          & 0.843 & 73.6 & 8.8  \\
  & $k = 400$          & 0.863 & 75.3 & 7.1  \\
  \bottomrule
\end{tabular}
\caption{A comparison of the side-view bending metrics (NMDM, MDM, BE) between
the biological baseline and simulated fillet meshes.}
\label{tab:sv_results}
\end{table}

\begin{figure*}[t]
\centering
\setlength{\tabcolsep}{2pt}
\renewcommand{\arraystretch}{0.5}
\begin{tabular}{c ccc c ccc}
  & \multicolumn{3}{c}{\textbf{Normal Fillet ($k = 0$)}} & & \multicolumn{3}{c}{\textbf{Severe WB fillet ($k = 100$)}} \\
  \cmidrule{2-4} \cmidrule{6-8}
      
  & Scene View & Top-Down View & Side View & \hspace{5pt} & Scene View & Top-Down View & Side View \vspace{4pt} \\

  \rotatebox[origin=c]{90}{} &
  \includegraphics[width=0.15\textwidth]{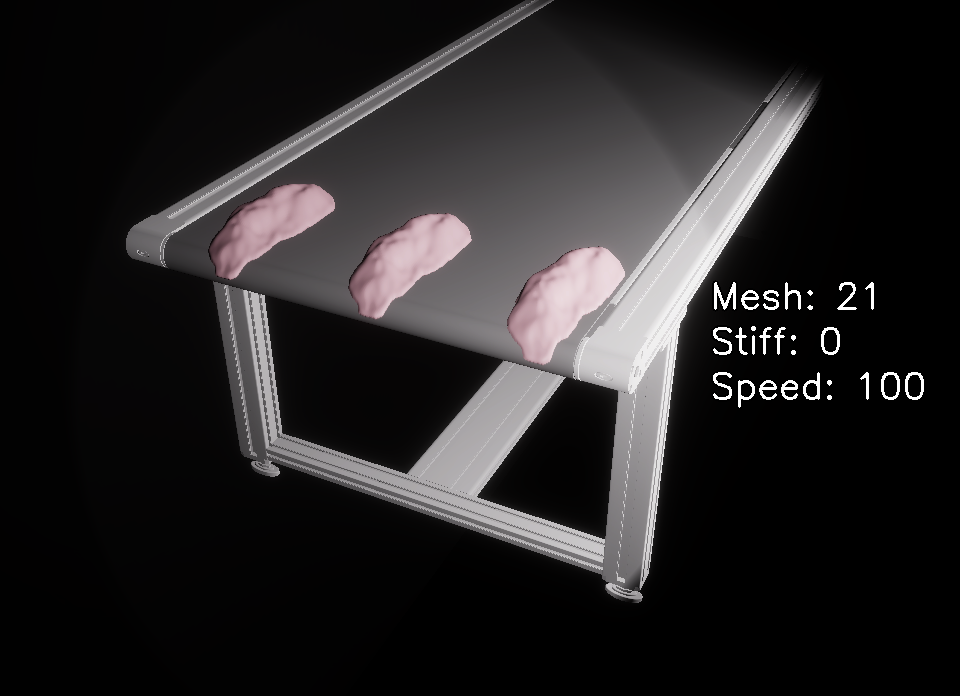} &
  \includegraphics[width=0.15\textwidth]{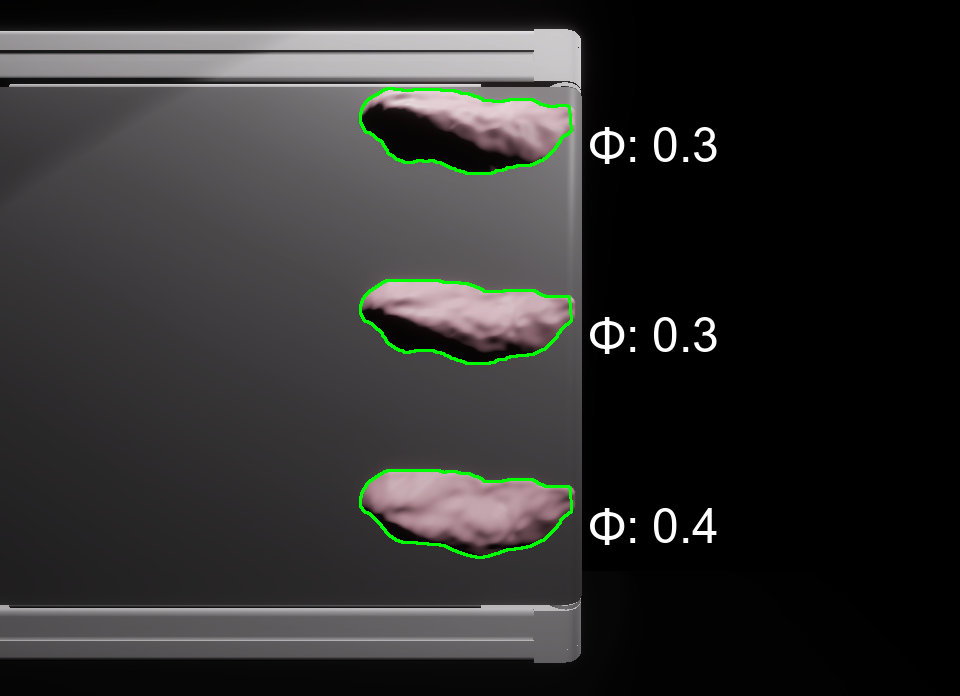} &
  \includegraphics[width=0.15\textwidth]{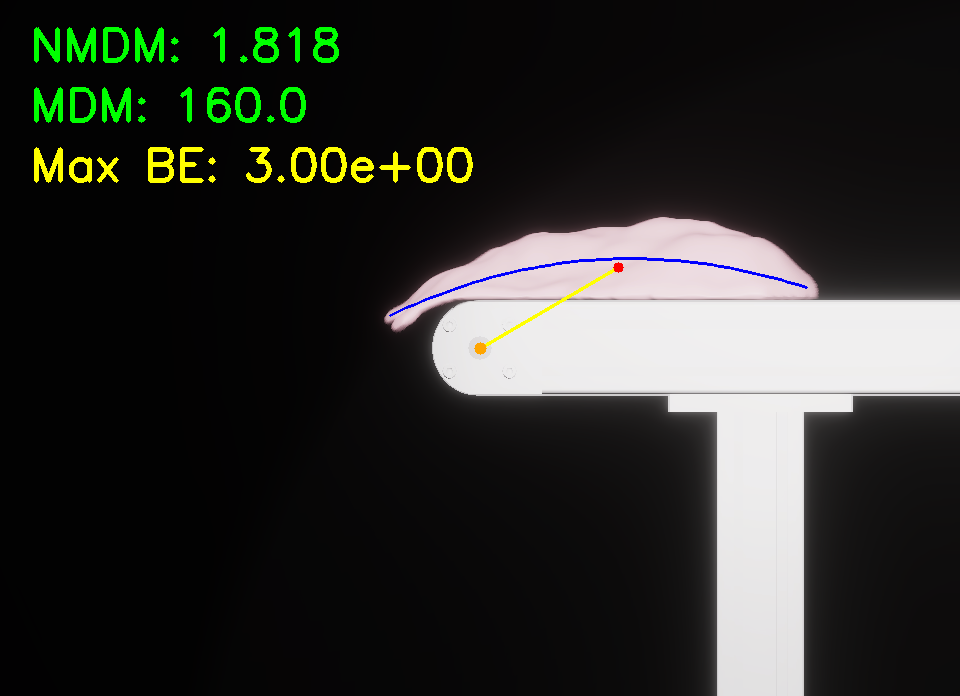} &
  &
  \includegraphics[width=0.15\textwidth]{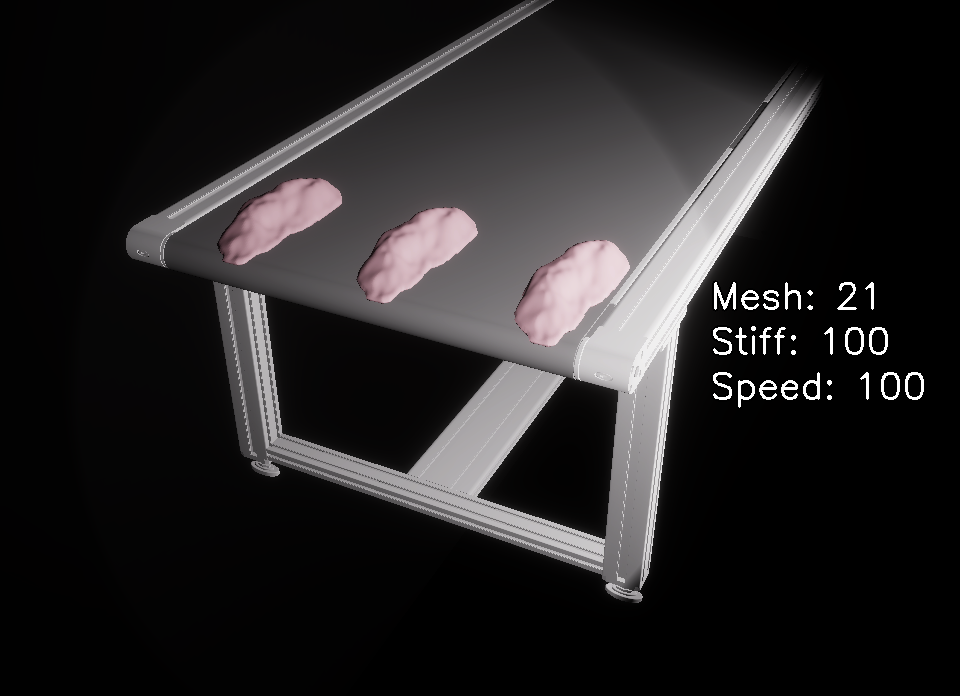} &
  \includegraphics[width=0.15\textwidth]{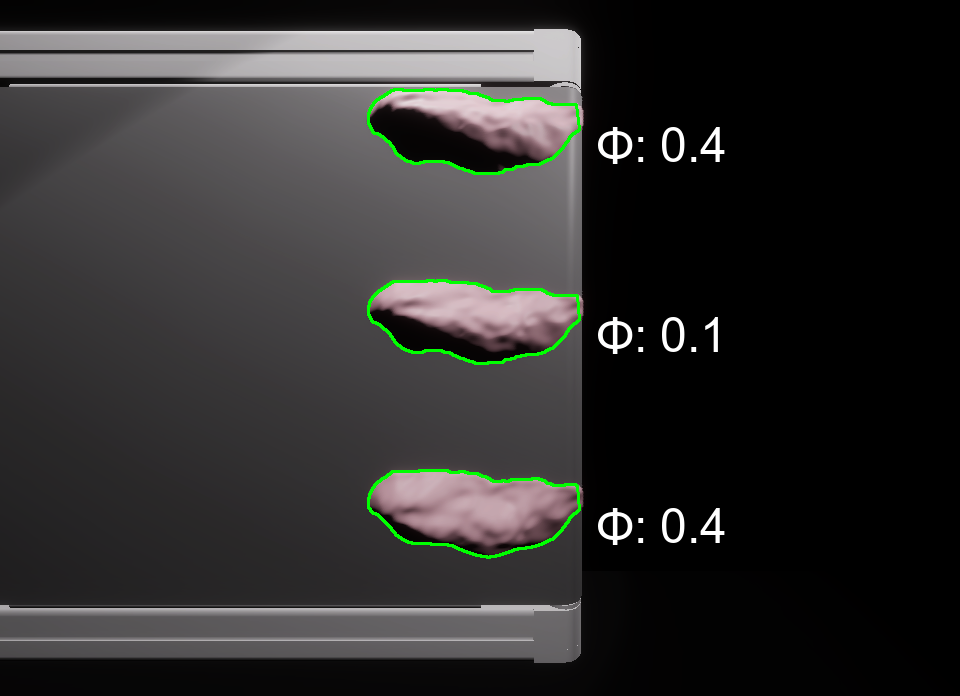} &
  \includegraphics[width=0.15\textwidth]{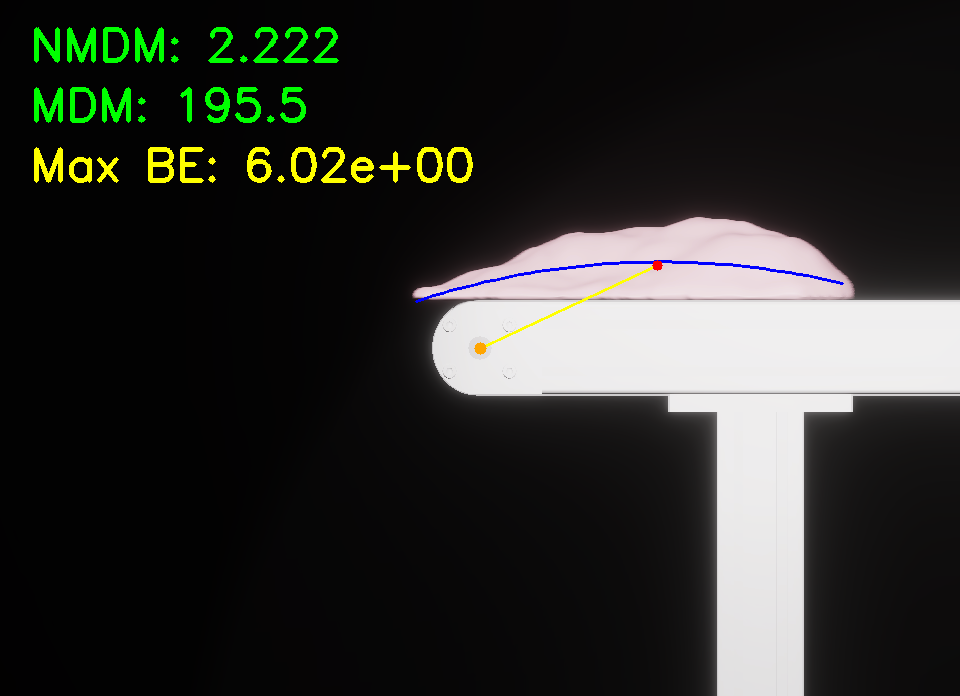} 
  \vspace{2pt} \\
  \rotatebox[origin=c]{90}{} &
  \includegraphics[width=0.15\textwidth]{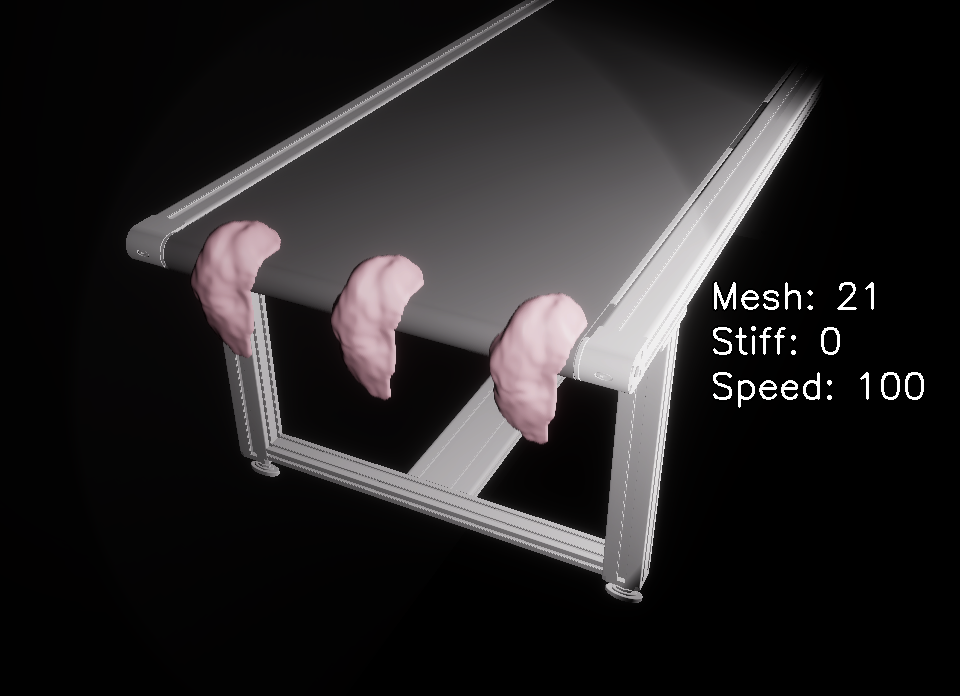} &
  \includegraphics[width=0.15\textwidth]{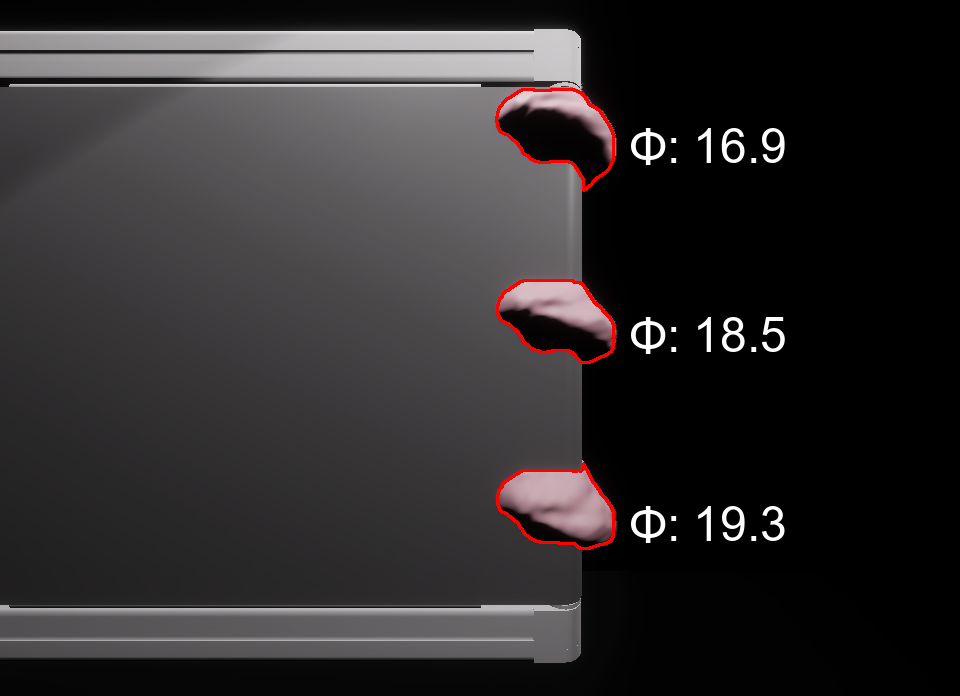} &
  \includegraphics[width=0.15\textwidth]{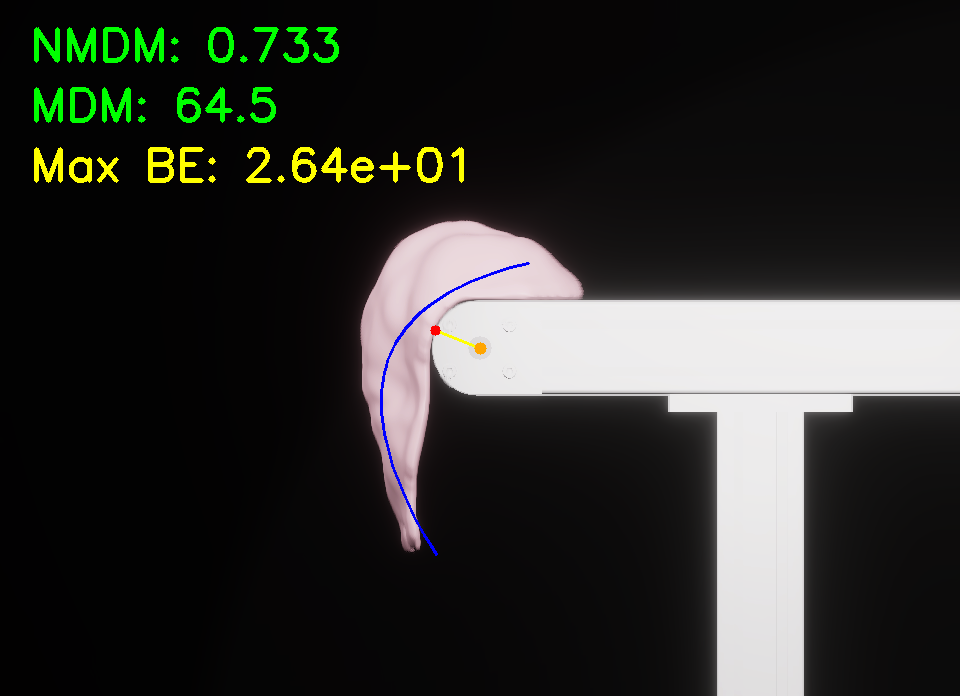} &
  &
  \includegraphics[width=0.15\textwidth]{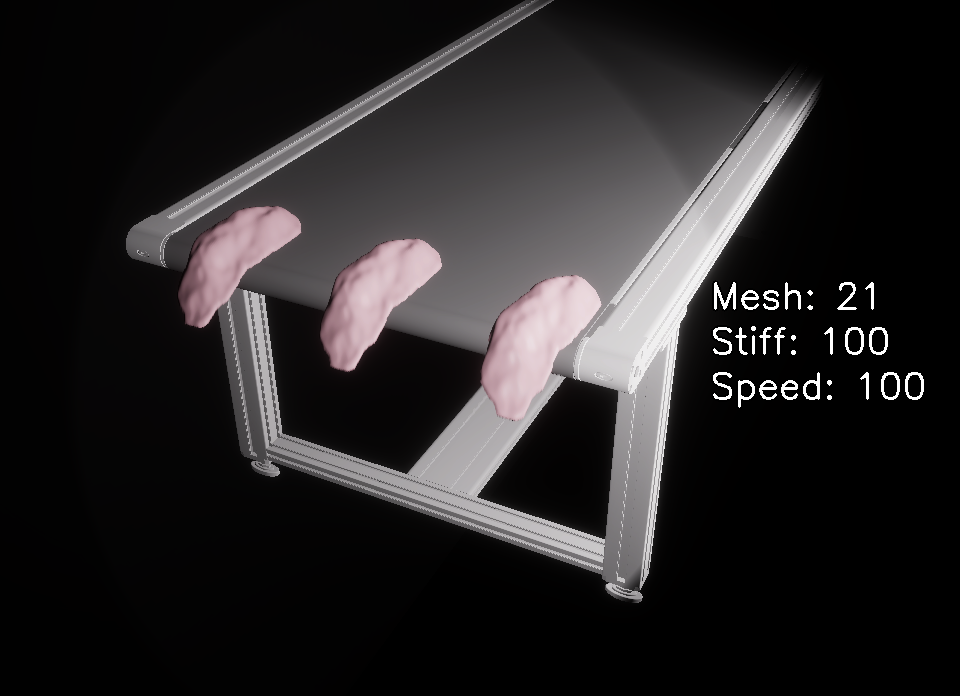} &
  \includegraphics[width=0.15\textwidth]{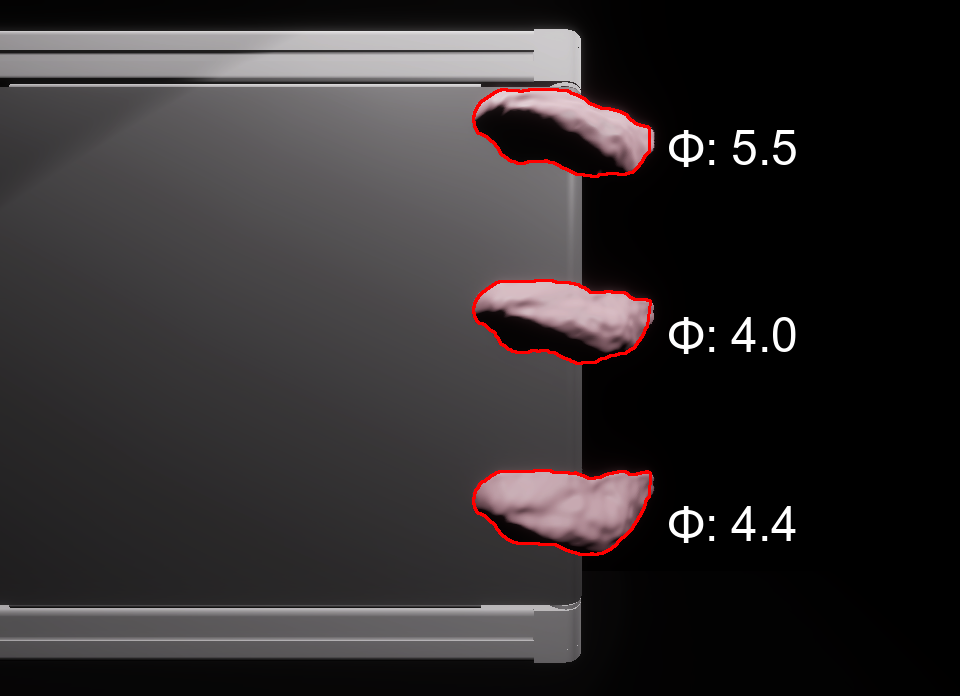} &
  \includegraphics[width=0.15\textwidth]{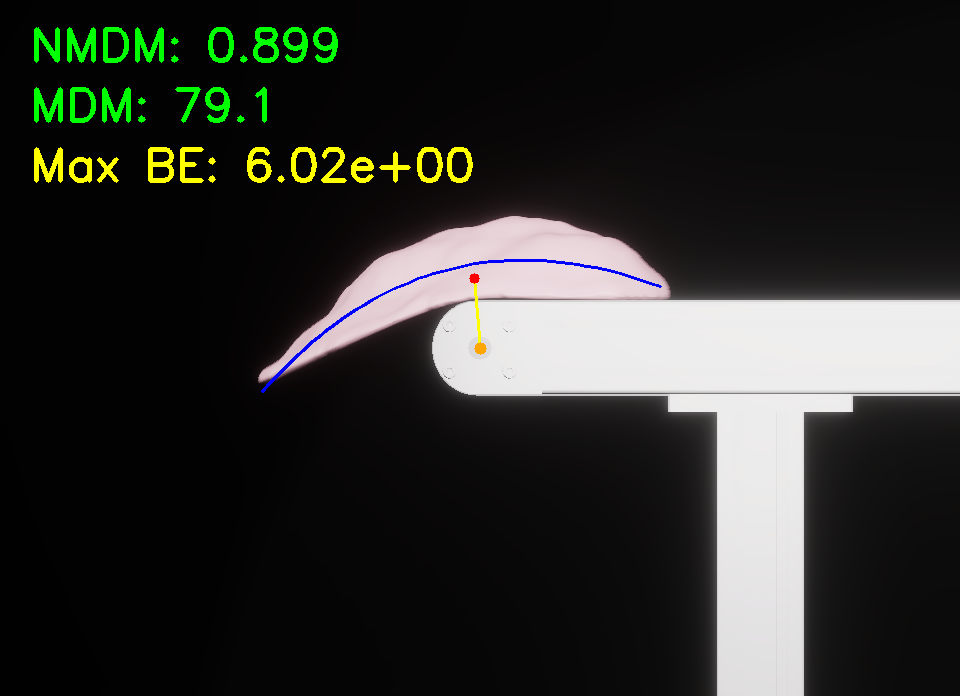} 
  \vspace{2pt} \\
  \rotatebox[origin=c]{90}{} &
  \includegraphics[width=0.15\textwidth]{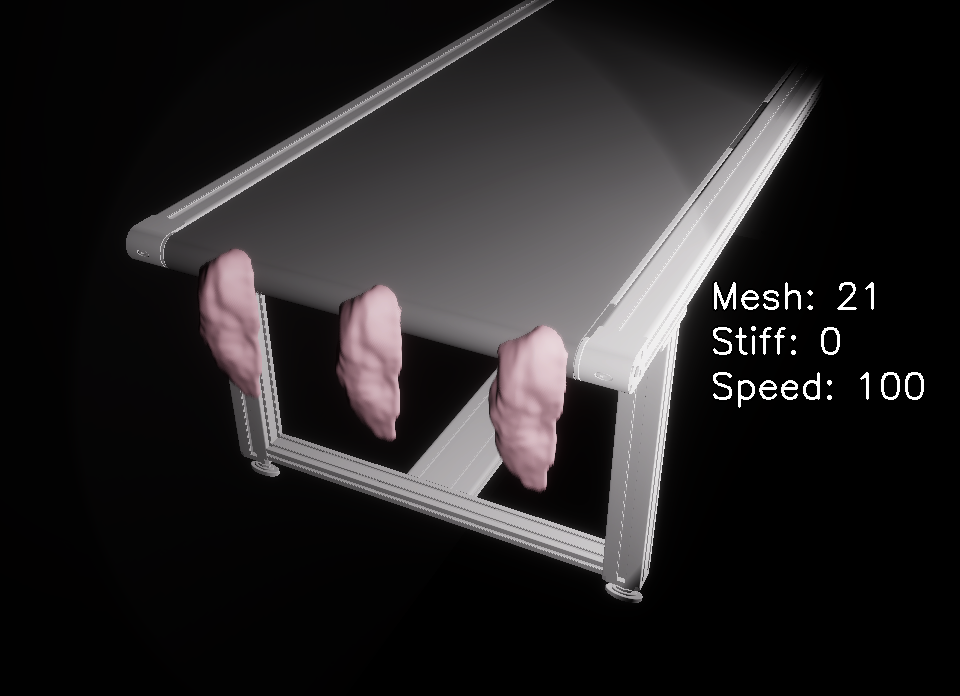} &
  \includegraphics[width=0.15\textwidth]{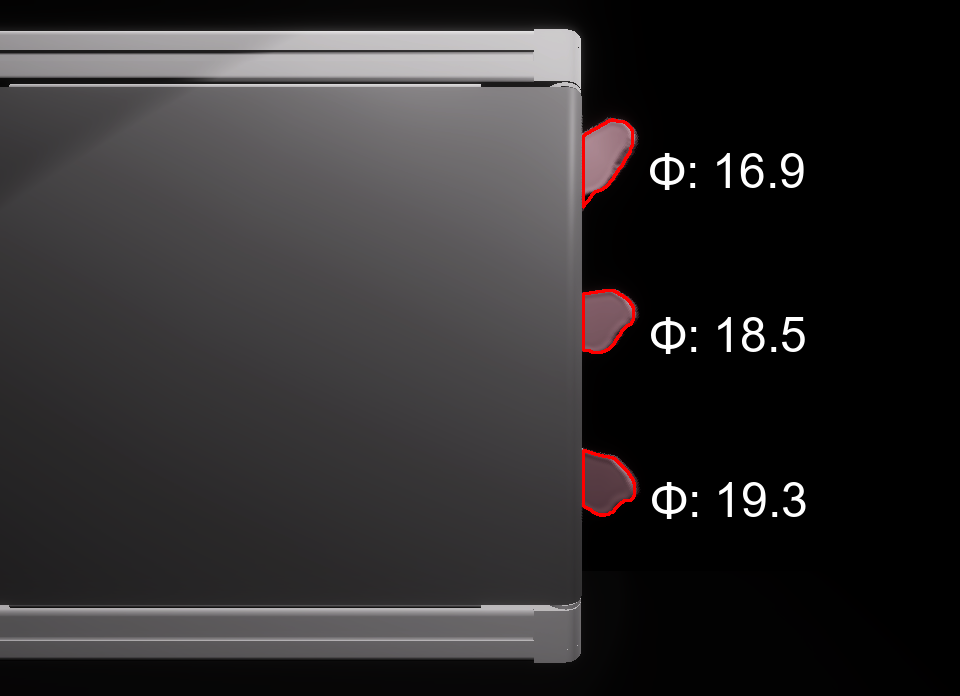} &
  \includegraphics[width=0.15\textwidth]{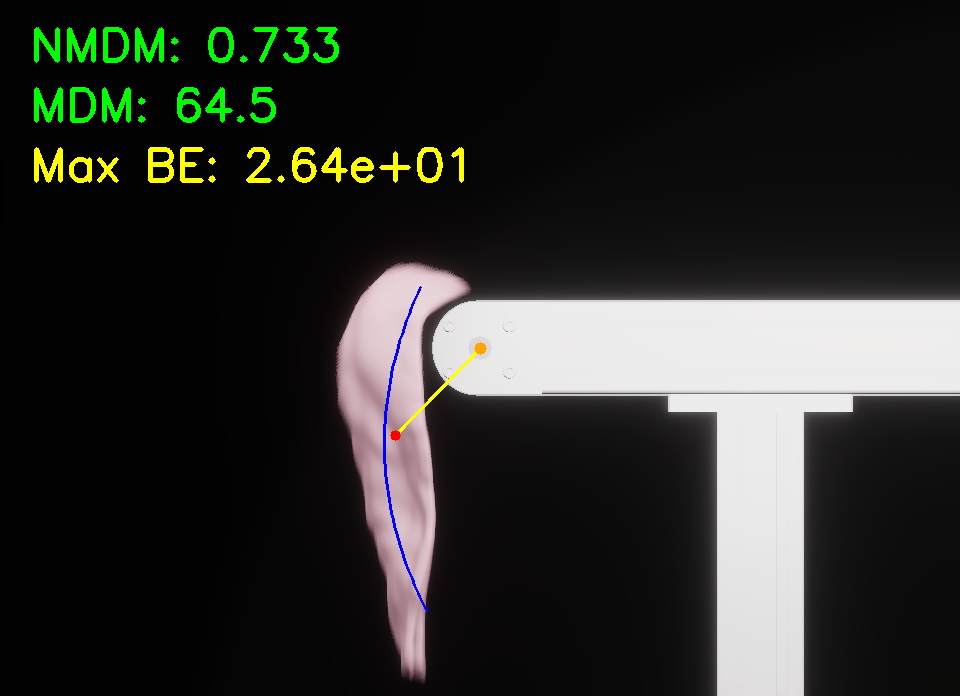} &
  &
  \includegraphics[width=0.15\textwidth]{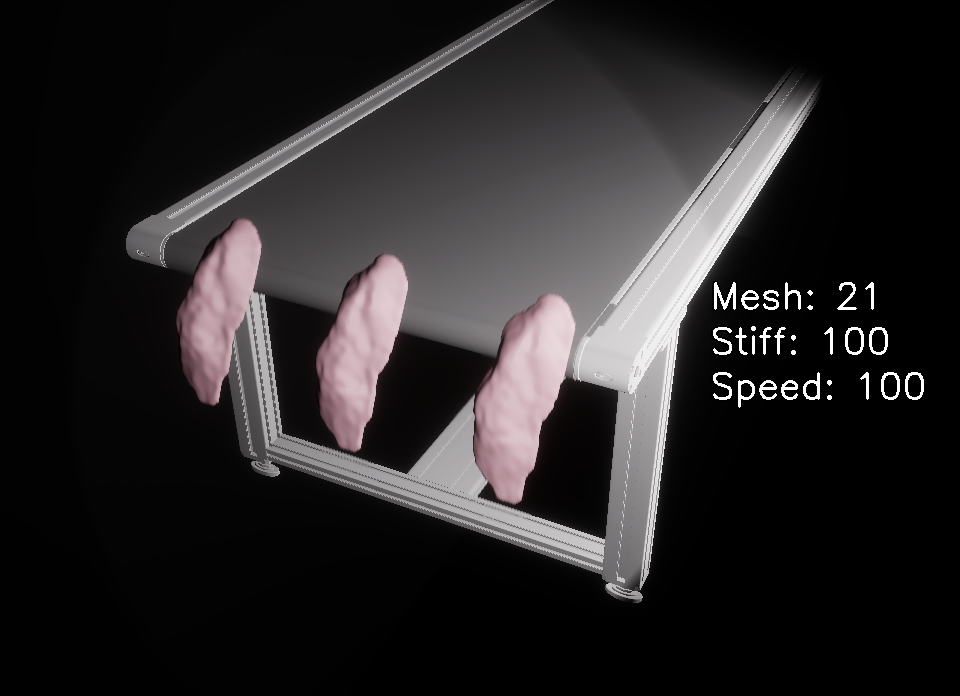} &
  \includegraphics[width=0.15\textwidth]{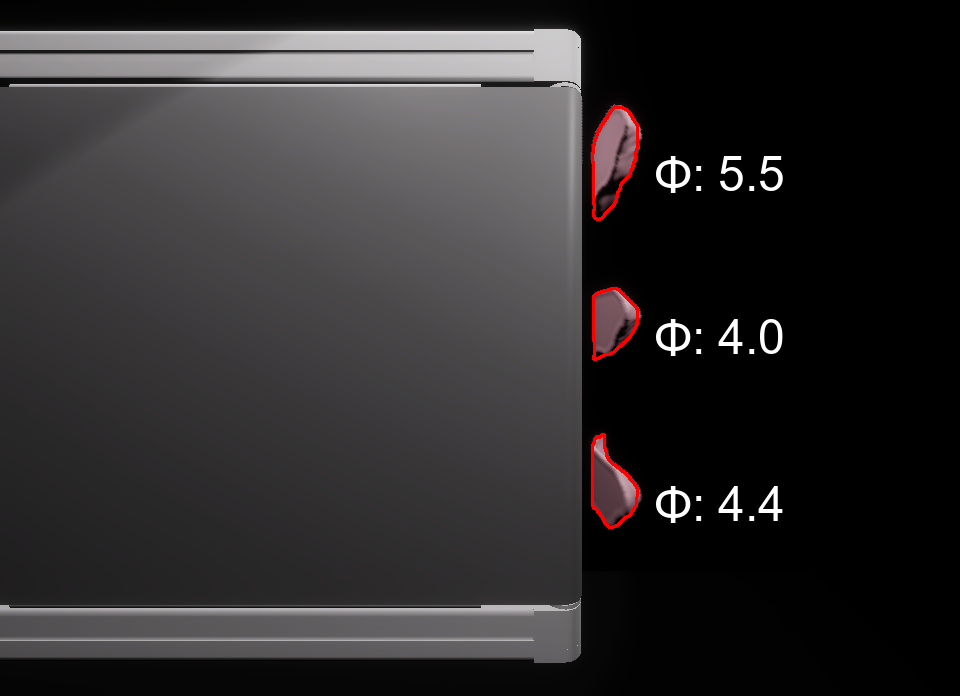} &
  \includegraphics[width=0.15\textwidth]{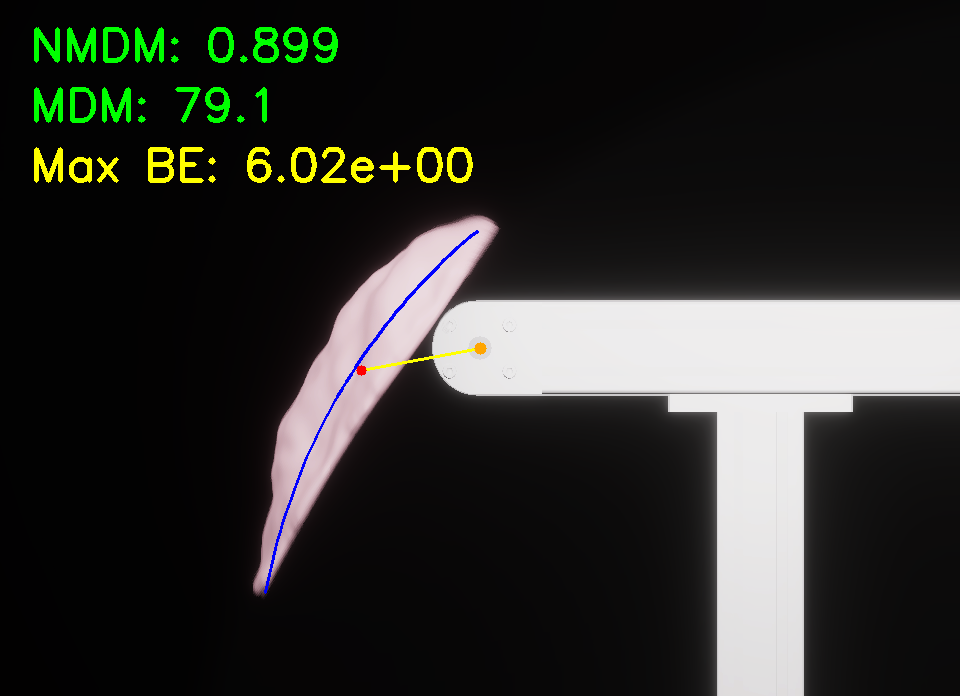} \hspace{-2pt}\\
\end{tabular}
\caption{A simulated visual comparison of a normal fillet ($k = 0$, left) and a
WB fillet ($k = 100$, right) traversing a conveyor belt at 100 FPM. The top row
illustrates the fillets fully supported on the belt, acting as the undeformed
baseline. In the middle row, the exact frame of the leading edge is captured as
it crosses the roller precipice, locking the maximum shape score. The bottom row
depicts the fillet in a free-fall state. A normal fillet exhibits high
compliance and structural deformation over the edge, whereas a severe WB fillet
maintains distinct rigidity.}
\label{fig:stiffness_comparison}
\vspace{-9pt}
\end{figure*}

We established $M=100$ as our working solution because its resulting kinematic
profile satisfies these boundary conditions as outlined in
Table~\ref{tab:sv_results}. At this mass, the distributed mass $m_i$ generates
the necessary downward torque to yield an NMDM of 0.520 for normal fillets at
the baseline $k=0$, and 0.783 for severe WB fillets at $k=100$. This produces a
simulated separation range of $\Delta\text{NMDM} = 0.263$, closely mirroring the
physical reality. Consequently, the combination of $M=100$ and $k \in [0, 100]$
was utilized for all subsequent multi-fillet experiments to represent the
progression of the WB myopathy. A simulated visual comparison of the bending
behavior between a normal fillet and a severe WB fillet traversing the conveyor
belt at 100 FPM is depicted in Fig.~\ref{fig:stiffness_comparison}.

\subsection{Top-Down View Camera Results}
The same collection of 1,000 synthetic fillet meshes ($M=100$) was subjected to
a drop test at 100 and 130 FPM. In contrast to the side-view system, which
assesses rigidity by measuring the change in the centroid of the fillet's
side-view profile, the top-down system quantifies the deformation of the
fillet's 2D silhouette as it bends over the roller edge. We employed the shape
score $\Phi$ as the primary metric for differentiation. As detailed in
Sec.~\ref{sec:methodology}, $\Phi$ aggregates the normalized change in Hu
moments and solidity. A highly flexible normal fillet ($k=0$) is expected to
exhibit significant contour deformation as the leading edge drapes over the
roller, resulting in a high $\Phi$ value. Conversely, a rigid WB fillet
($k=100$) should maintain its structure, yielding a minimal change in moments
and solidity, and thus a significantly lower $\Phi$ score.

\begin{table}
\centering
\setlength{\tabcolsep}{2.5pt} 
\renewcommand{\arraystretch}{1.1}
\begin{tabular}{cc ccc}
  \toprule
  \textbf{Speed} & \textbf{Stiffness ($k$)} & \textbf{Shape Score ($\Phi$)} & \textbf{Hu Mo. Score} & \textbf{Sol. Diff.} \\
  \midrule
  \multirow{6}{*}{100 FPM} 
  & 0.0 (normal)   & 19.12 $\pm$ 5.67 & 0.178 & 0.014 \\
  & 20.0           & 11.16 $\pm$ 4.42 & 0.104 & 0.008 \\
  & 40.0           & 8.79 $\pm$ 3.65  & 0.082 & 0.007 \\
  & 60.0           & 7.09 $\pm$ 3.10  & 0.066 & 0.006 \\
  & 80.0           & 6.28 $\pm$ 2.83  & 0.058 & 0.005 \\
  & 100.0 (severe) & 5.40 $\pm$ 2.54  & 0.050 & 0.005 \\
  \midrule
  \multirow{6}{*}{130 FPM}
  & 0.0 (normal)   & 14.91 $\pm$ 5.44 & 0.139 & 0.011 \\
  & 20.0           & 7.27 $\pm$ 3.56  & 0.067 & 0.006 \\
  & 40.0           & 5.88 $\pm$ 3.22  & 0.054 & 0.005 \\
  & 60.0           & 4.87 $\pm$ 2.82  & 0.045 & 0.005 \\
  & 80.0           & 4.25 $\pm$ 2.28  & 0.039 & 0.004 \\
  & 100.0 (severe) & 4.01 $\pm$ 3.03  & 0.036 & 0.004 \\
  \bottomrule
\end{tabular}
\caption{Top-down shape scores across line speeds.}
\label{tab:tdv_results}
\end{table}

The simulation results confirm that the top-down shape score serves as a highly
robust discriminator for the WB condition. As summarized in
Table~\ref{tab:tdv_results}, at a line speed of 100 FPM, normal fillets ($k=0$)
exhibited a mean shape score of $19.12 \pm 5.67$, indicating substantial
geometric deformation. In contrast, severe WB fillets ($k=100$) yielded a mean
score of only $5.40 \pm 2.54$. This represents a separation factor of nearly
$3.5\times$, demonstrating that the simulated rigidity effectively suppresses
the contour distortion observed in normal fillets. Increasing the line speed to
130 FPM reduced the absolute magnitude of the deformation for all classes. This
reduction is driven by the increased forward momentum of the fillets. However,
despite this shift the relative separation remained distinct. Normal fillets
scored $14.91 \pm 5.44$ while severe WB fillets scored $4.01 \pm 3.03$.

\begin{table}
\centering
\setlength{\tabcolsep}{5pt}
\renewcommand{\arraystretch}{1.1}
\begin{tabular}{cccc}
  \toprule
  \textbf{Lane Position} & \textbf{Normal ($k=0$)} & \textbf{Severe ($k=100$)} & \textbf{Sep. ($\Delta\Phi$)} \\
  \midrule
  Lane 0 (outer)  & 19.06 $\pm$ 5.53 & 5.28 $\pm$ 2.56 & 13.78 \\
  Lane 1 (center) & 19.34 $\pm$ 5.81 & 5.14 $\pm$ 2.49 & 14.20 \\
  Lane 2 (outer)  & 18.97 $\pm$ 5.66 & 5.78 $\pm$ 2.54 & 13.20 \\
  \bottomrule
\end{tabular}
\caption{Shape score lane analysis. The center lane (1) captures the maximum
geometric deformation while the outer lanes (0 and 2) exhibit slightly lower
range scores}
\label{tab:lane_analysis}
\end{table}

A key challenge in single-camera multi-fillet tracking is perspective
distortion. Fillets positioned at the edges of the conveyor belt are viewed at
an oblique angle compared to those in the center lane. We analyzed the mean
shape score across the three distinct conveyor lanes at 100 FPM, as shown in
Table~\ref{tab:lane_analysis}. A slight perspective bias was observed. The
center lane (lane 1) yielded the highest range ($\Delta \Phi = 14.20$). The
outer lanes (lanes 0 and 2) exhibited a marginally compressed range ($\Delta
\Phi \approx 13.5$). Despite this variance, the separation between normal and WB
samples remained statistically robust across all lanes. The lowest mean score
for a normal fillet in an outer lane ($\Phi = 18.97$) was still over $3\times$
higher than the highest mean score for a WB fillet ($\Phi = 5.78$).

\section{Conclusion and Future Work}
\label{sec:conclusion_and_future_work}
This paper presented a novel simulation framework to address the throughput
limitations of single-fillet WB detection in industrial poultry processing
plants. By generating a synthetic dataset of 3D fillet meshes and modeling their
dynamic bending behavior over a conveyor roller, we demonstrated that
multi-fillet evaluation is achievable using a single top-down camera.
Furthermore, our experimental evaluation indicates that a top-down shape score
quantifying 2D contour deformation shows strong potential as a WB discriminator.
Notably, simulated normal fillets exhibited significant geometric deformation,
yielding a separation factor of nearly $3.5\times$ compared to severe WB fillets
at a line speed of 100 FPM. For future work, we plan to transition the
multi-fillet tracking system from the simulated environment to a physical,
real-world conveyor processing line. 


\section*{Acknowledgments}
\label{sec:acknowledgment}
This material is based upon work supported by the United States Department of
Agriculture (USDA) under Agricultural Research Service CRIS Project
\#6066-21310-006-000-D.

\bibliographystyle{IEEEtran}
\bibliography{simulation-based_multi-fillet_evaluation_of_woody_breast_poultry_fillets}

\end{document}